\pdfoutput=1
\documentclass[11pt]{article}
\usepackage[final]{acl}
\usepackage{times}
\usepackage{latexsym}
\usepackage[T1]{fontenc}
\usepackage[utf8]{inputenc}
\usepackage{microtype}
\usepackage{inconsolata}
\usepackage{graphicx}
\usepackage{booktabs}
\usepackage{fancyvrb}
\usepackage{xspace}
\usepackage{pifont}
\usepackage{hyperref}
\usepackage{listings} 
\usepackage{amsmath,amsfonts,amssymb} 
\usepackage{enumitem}
\usepackage{tabu}
\usepackage{adjustbox}
\usepackage{tabularx, booktabs}
\usepackage{multirow}
\usepackage{algorithm}
\usepackage{algorithmicx}
\usepackage{algpseudocode}

\newcommand{\llm}{\textsc{Llm}\xspace} 

\newcommand{\daily}{\textsc{DailyDialog}\xspace}
\newcommand{\seedd}{\emph{Seed Dialogue}\xspace}
\newcommand{\seeddp}{\emph{Seed Dialogue Pool}\xspace}
\newcommand{\mathsdp}{\mathbf{D}_{seed}}

\newcommand{\seedds}{\emph{Seed Dialogue Summary}\xspace}
\newcommand{\seeddsp}{\emph{Seed Dialogue Summary Pool}\xspace}
\newcommand{\mathsdsp}{\mathbf{S}_{seed}}

\newcommand{\adsp}{\emph{Model-generated Dialogue Summary Pool}\xspace}
\newcommand{\mathadsp}{\mathbf{S}_{aug}}
\newcommand{\adp}{\emph{Model-generated Dialogue Pool}\xspace}
\newcommand{\mathadp}{\mathbf{D}_{aug}}

\title{Controllable and Diverse Data Augmentation  with Large Language Model for Low-Resource Open-Domain Dialogue Generation}

\author{
 \textbf{Zhenhua Liu},
 \textbf{Tong Zhu},
 \textbf{Jianxiang Xiang},
 \textbf{Wenliang Chen}
 \\
 Soochow University
 \\
 \texttt{\{zhliu0106, tzhu7, jxxiang0720\}@stu.suda.edu.cn}
 \\
 \texttt{wlchen@suda.edu.cn}
}

\begin{document}
\maketitle
\begin{abstract}
Data augmentation (DA) is crucial to mitigate model training instability and over-fitting problems in low-resource open-domain dialogue generation.
However, traditional DA methods often neglect semantic data diversity, restricting the overall quality.
Recently, large language models (\llm) have been used for DA to generate diversified dialogues. However, they have limited controllability and tend to generate dialogues with a distribution shift compared to the seed dialogues.
To maximize the augmentation diversity and address the controllability problem, we propose \textbf{S}ummary-based \textbf{D}ialogue \textbf{A}ugmentation with \llm (SDA).
Our approach enhances the controllability of \llm by using dialogue summaries as a planning tool.
Based on summaries, SDA can generate high-quality and diverse dialogue data even with a small seed dataset.
To evaluate the efficacy of data augmentation methods for open-domain dialogue, we designed a clustering-based metric to characterize the semantic diversity of the augmented dialogue data.
The experimental results show that SDA can augment high-quality and semantically diverse dialogues given a small seed dataset and an \llm, and the augmented data can boost the performance of open-domain dialogue models.
\end{abstract}

\section{Introduction}

Data-driven deep learning models often require large amounts of data, which is especially important for open-domain dialogue generation \cite{zhang2019dialogpt, Roller2020RecipesFB}. 
However, data resources are usually scarce in new dialogue scenarios (like counseling or empathetic dialogues).
Furthermore, it is difficult to annotate dialogues given the context, since there are multiple plausible responses.
As a result, the collection of large amounts of high-quality and semantically diverse dialogue data is extremely expensive and time-consuming \cite{Li2017DailyDialogAM, Zhang2018PersonalizingDA, Dinan2018WizardOW}.

\begin{figure}[t]
    \centering
    \includegraphics[width=\linewidth]{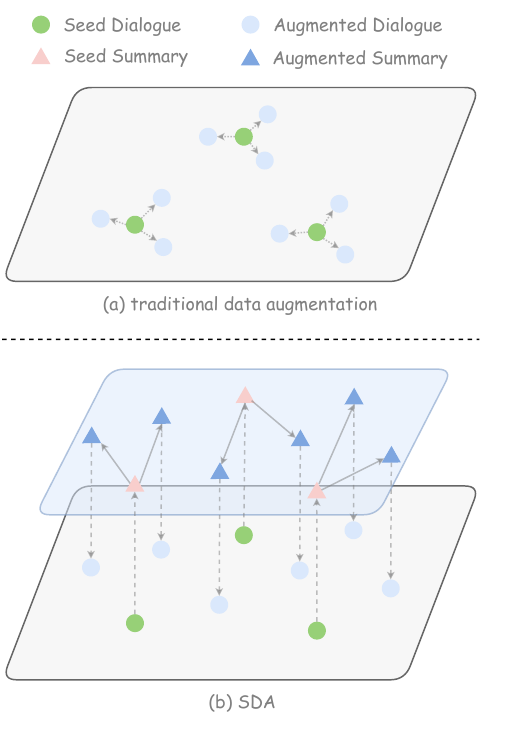}
    \caption{Overview of traditional data augmentation method and Summary-based Dialogue Augmentation with \llm (SDA).}
    \label{fig:intro}
\end{figure}

A feasible solution is data augmentation (DA) \cite{Shorten2019ASO, feng2021survey}, but it struggles to perform high-quality augmentation when the seed dataset is small.
Traditional DA methods for natural language processing include rule-based methods \cite{Xie2019UnsupervisedDA, Wei2019EDAED, Karimi2021AEDAAE} and model-based methods \cite{Sennrich2015ImprovingNM, Yang2020GDAugGD, Ng2020SSMBASM, AnabyTavor2020DoNH}, which limit the data diversity \cite{Xu2023WizardLMEL}. As shown in \autoref{fig:intro} (a), the traditional DA methods usually perturb the seed data at word-level or sentence-level, with little semantic differences. 
Specifically, several approaches of data augmentation have been proposed for open-domain dialogue systems and achieved a certain success \cite{Zhang2020DialogueDO, Ou2022CounterfactualDA}. However, the semantic diversity of the augmented dialogues generated by these methods is still constrained by the \seedd. In addition, these methods are difficult to apply to low-resource scenarios.

Recently, \llm has shown great potential in various natural language processing tasks with in-context learning (ICL). Given an instruction and a few exemplars, \llm can perform a series of complex tasks \cite{Brown2020LanguageMA, Dong2022ASF}. 
In this paper, we investigate the augmentation of a small \seedd dataset using solely the \llm.
Since it is trained on huge data, \llm can provide diversity for our task. 
Nevertheless, directly prompting the \llm usually lacks controllability and tends to generate dialogues with a distribution shift compared to the \seedd.

Based on the above challenges and problems, we propose \textbf{S}ummary-based \textbf{D}ialogue \textbf{A}ugmentation with \llm (SDA) for low-resource open-domain dialogue generation.
A three-step approach as shown in \autoref{fig:intro} (b): Firstly, we convert the \seedd into \seedds with the assistance of \llm, which briefly summarizes the main topics and contents of the dialogue. Secondly, we leverage the \seedds to generate more dialogue summaries with a wide diversity of topics. Finally, we take the augmented dialogue summaries as the planning to generate a dialogue. Directly prompting \llm usually lacks controllability and tends to generate unexpected dialogue. Our solution takes a different way by taking \emph{Dialogue Summary} as the planning to prompt \llm. The dialogue summary, as an abstract representation of a dialogue, can briefly present the main topics and contents of the dialogue, which improves \llm's controllability. In the end, we could obtain a \adp which contains a large amount of dialogue data both high-quality and diverse with a similar distribution as the \seedd.

To evaluate the efficacy of our proposed method, we design a clustering-based metric, \textsc{SemanticDiversity}, to characterize the diversity of the augmented dialogue on the basis of the same distribution as the \seedd. Unlike metrics such as Distinct \cite{Li2015ADO}, which evaluate data diversity at the word-level, \textsc{SemanticDiversity} can evaluate the diversity of augmented dialogues at the semantic-level. 
Experimental results indicate that, given a small seed dataset and \llm, SDA can effectively augment dialogues with both high quality and semantic diversity. Furthermore, this augmented data enhances the performance of open-domain dialogue generation models.

In summary, our contributions include: 

\begin{itemize}
\item[$\bullet$] We propose SDA, a dialogue augmentation approach that exploits the potential of \llm to augment the given small seed data. Our approach improves the controllability of \llm by using summary as planning, which generates high-quality and diverse dialogue data that matches the distribution of seed data. 
\end{itemize}

\begin{itemize}
\item[$\bullet$] We develop a new clustering-based metric \textsc{SemanticDiversity}, which could characterize the semantic diversity of the augmented dialogue on the basis of the same distribution as the \seedd. We conduct a comprehensive analysis of augmented dialogue data to demonstrate the superior data quality and diversity generated by SDA, compared to other baseline methods.
\end{itemize}

\begin{itemize}
\item[$\bullet$] Extensive experiments show that our proposed solution can boost model performance in low-resource scenarios.
\end{itemize}

\begin{figure*}
    \centering
    \includegraphics[width=\linewidth]{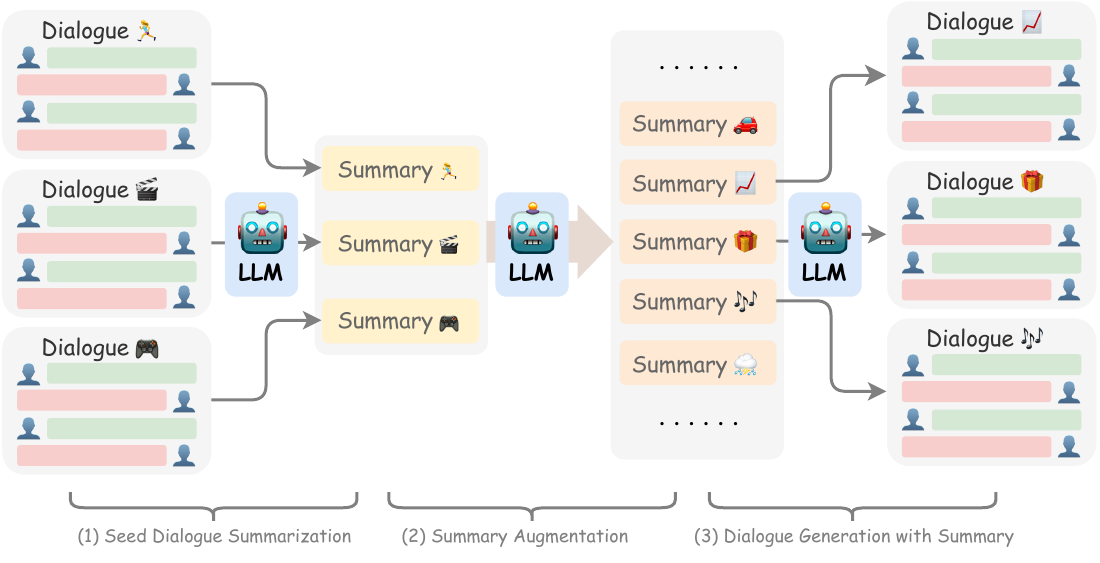}
    \caption{Main framework of our method. The process is in three steps: (1) First, summarizing the seed dialogue into dialogue summary. (2) Secondly,  we leverage the seed dialogue summary to generate more dialogue summaries with a wide diversity of topics. (3) Finally, we convert the augmented dialogue summary back into dialogue. All these steps are performed by \llm.}
    \label{fig:main_framework}
\end{figure*}

\section{Related Work}

Related work involves in-context learning and data augmentation.

\noindent\textbf{In-Context Learning.} With the increasing ability of large language models, 
 in-context learning (ICL) performs few-shot learning by doing inference conditioning on several exemplars \cite{Brown2020LanguageMA, Dong2022ASF}. ICL has obtained success in semantic parsing \cite{Pasupat2021ControllableSP, Rubin2021LearningTR, Shin2021FewShotSP}, information extraction \cite{Wan2023GPTREIL, He2023ICLD3IEIL}, machine translation \cite{Zhu2023MultilingualMT, Sia2023IncontextLA}, and other Natural Language Processing tasks. In particular, there have been some previous attempts to apply ICL to dialogue systems \cite{Yu2021FewshotIC, Parikh2023ExploringZA, Xie2022UnifiedSKGUA, Hu2022InContextLF, Madotto2021FewShotBP}. In this paper, we use the capabilities of ICL to perform data augmentation on small seed dialogue data.

\medskip

\noindent\textbf{Data Augmentation.} The traditional data augmentation methods for natural language processing include rule-based methods \cite{Xie2019UnsupervisedDA, Wei2019EDAED, Karimi2021AEDAAE} and model-based methods \cite{Sennrich2015ImprovingNM, Yang2020GDAugGD, Ng2020SSMBASM, AnabyTavor2020DoNH}. As shown in \autoref{fig:intro}, the augmented data obtained using these methods are usually word-level or sentence-level alternative representations of the seed data, with little semantic differences. In addition to the traditional methods, another line of work is prompting \llm to augment various natural language processing datasets \cite{Chen2022WeaklySD, Wang2022PromDAPD, Sahu2022DataAF, Mehri2022LADLM, Rosenbaum2022CLASPFC}. In particular, several approaches have been proposed for open-domain dialogue systems \cite{Zhang2020DialogueDO, Ou2022CounterfactualDA}. However, the semantic diversity of the augmented dialogues generated by these methods is still constrained by the seed data and the data distribution does not necessarily match the seed data distribution. In addition, these methods are difficult to apply to low-resource scenarios. Instead, our method improves the controllability of \llm, which can generate high-quality and diverse dialogues that match the distribution of seed data.
\section{Methodology}

Conventional techniques for data augmentation frequently produce dull and monotonous dialogue content that lacks diversity. To remedy this problem, we propose SDA, \textbf{S}ummary-based \textbf{D}ialogue \textbf{A}ugmentation, where all these procedures are accomplished through \llm's In-Context Learning (ICL). With the assistance of ICL, \llm can accomplish the specified task without fine-tuning model parameters. In this paper, we choose LLaMA-7B \cite{touvron2023llama} as the backbone, while it is possible to apply our approach with other LLMs. The main framework is illustrated in \autoref{fig:main_framework}, which includes Seed Dialogue Summarization, Dialogue Summary Augmentation, and Dialogue Generation with Summary.


\subsection{Task definition}

\label{sec:task_difinition}

The \seeddp, $\mathsdp = \{d_1, d_2, ..., d_n\}$, consists of $n$ dialogues, where dialogue $d=\{u_1, u_2, ... u_k\}$ includes $k$ utterances. With a given \llm $M$ and $\mathsdp$, our objective is to obtain a \adp $\mathadp$ consisting of $m$ dialogues that are both high-quality and diverse, where $m\gg n$.

\begin{table*}[ht]
    \centering
    \footnotesize
    \begin{Verbatim}[frame=single]
Summarize the following dialogue between User A and User B:
Exemplar 1:
User A: Hello, this is Lucy. May I speak to Mr. Smith?
User B: Oh, hello, Lucy. What's up?
User A: I'm afraid I can't come to work today, Mr. Smith.
User B: Oh, what's wrong with you?
Summary:
In the above dialogue, User A calls User B and asks to speak to Mr. Smith. User B answers and 
they exchange greetings. User A informs User B that he won't be able to come to work today ...
...
Exemplar 6:
[Seed Dialogue]
Summary:
In the above dialogue,
\end{Verbatim}
    \caption{Prompt used for summarizing the seed dialogue. Due to the space limit, we only display one exemplar in the table.}
    \label{tab:prompt_d2s}
\end{table*}

\begin{table*}[ht]
    \centering
    \footnotesize
    \begin{Verbatim}[frame=single]
Two people are chatting with each other, here are some possible summaries of their dialogue:
Summary 1: [summary 1 from seed dialogue summary pool]
Summary 2: [summary 2 from seed dialogue summary pool]
Summary 3: [summary 3 from seed dialogue summary pool]
Summary 4: [summary 4 from seed dialogue summary pool]
Summary 5: [summary 5 from seed dialogue summary pool]
Summary 6: [summary 6 from model-generated dialogue summary pool]
Summary 7: [summary 7 from model-generated dialogue summary pool]
Summary 8: [summary 8 from model-generated dialogue summary pool]
Summary 9: 
\end{Verbatim}
    \caption{Prompt used for generating new summaries.}
    \label{tab:prompt_s2s}
\end{table*}

% ...

\begin{table*}[ht]
    \centering
    \footnotesize
    \begin{Verbatim}[frame=single]
Convert the following summary into dialogue:
Exemplar 1:
In the above dialogue, User A calls User B and asks to speak to Mr. Smith ...
Dialogue:
User A: Hello, this is Lucy. May I speak to Mr. Smith?
User B: Oh, hello, Lucy. What's up?
User A: I'm afraid I can't come to work today, Mr. Smith.
User B: Oh, what's wrong with you?
...
Exemplar 6:
[Dialogue Summary]
Dialogue:
User A:
\end{Verbatim}
    \caption{Prompt used for turning the summary into a dialogue. Due to the space limit, we only display one exemplar in the table.}
    \label{tab:prompt_s2d}
\end{table*}

\subsection{Seed Dialogue Summarization}
\label{sec:summarization}

To generate diverse and informative dialogues, we first summarize every seed dialogue data $d$ into a seed dialogue summary $s$. The dialogue summary, as an abstract representation of a dialogue, can briefly present the main topics and contents of the dialogue. We write a prompt $p_{d2s}$, accompanied by a task description and 5 exemplars, to improve the performance of ICL, which can be found in \autoref{tab:prompt_d2s}.

Given $p_{d2s}$ and $d \in \mathsdp$, we can obtain the dialogue summary $s$ with \llm $\mathbf{M} $: $$s = \mathbf{M}(p_{d2s}, d).$$

Afterwards, we obtain the \seeddsp $\mathsdsp = \{s_1, s_2, ..., s_n\}$, which contains $n$ dialogue summaries corresponding to the seed dialogues.

\subsection{Dialogue Summary Augmentation}
\label{sec:dialogue_summary_generation}

\llm can be prompted to generate new and novel dialogue summaries when presented with some existing summaries. In this way, we can augment summaries from a small set of seed data. Given the \seeddsp $\mathsdsp$, we propose a method in the bootstrapping fashion to generate diverse dialogue summaries. For every step, we sample 8 dialogue summaries as in-context exemplars and then prompt the \llm to generate a new dialogue summary. Of the 8 exemplar dialogue summaries, 5 are sampled from \seedds, and 3 are from \adsp to promote diversity. The new dialogue summary is then added to the \adsp $\mathadsp$. It is worth noting that when $\mathadsp$ is empty, all the 8 exemplars are sampled from the $\mathsdsp$. The procedure repeats until $\mathadsp$ reaches a certain size $m$. The prompt $p_{s2s}$ is shown in \autoref{tab:prompt_s2s}.

\subsection{Dialogue Generation with Summary}
\label{sec:dialogue_data_generation}

Next, we take dialogue summary $s \in \mathadsp$ as the planning to generate a dialogue $d_{new}$. To improve the controllability and quality, the summary $s$ is used as planning when generating the new dialogue $d_{new}$. The summary contains the main topics and contents of the conversation. Based on this, we can prompt \llm to generate the dialogue data. The prompt $p_{s2d}$ is shown in \autoref{tab:prompt_s2d}.

As mentioned in \autoref{sec:task_difinition}, a dialogue $d=\{u_1, u_2, ... u_k\}$ consist of $k$ utterances. So we need to iteratively generate utterance $u_i$ based on the dialogue summary $s$ and the previously generated $u_1, ..., u_{i-1}$. Given $p_{s2d}$, $s \in \mathadsp$ and $u_1, ..., u_{i-1}$, we can obtain the utterance $u_i$ with \llm $\mathbf{M} $: $$u_i = \mathbf{M}(e_{s2d}, s, u_1, ..., u_{i-1}).$$

We repeat this process until the number of utterances in a dialogue data $d$ is greater than 3 and the last utterance contains 'bye' or 'see you'.
In the end, we obtain the final \adp $\mathadp = \{d_1, d_2, ..., d_m\}$, which contains $m$ model-generated dialogue data.

\subsection{Data Filtering}

\label{sec:data_filtering}

The limitations of the \llm's capabilities may result in unsatisfactory model-generated dialogue summaries or dialogue data. As a result, filtering the generated data becomes necessary.

\textbf{Summary Filtering.} We only keep the dialogue summaries that include `User A' and `User B', and make sure the length of summary has at least 18 tokens. In order to enhance diversity, we compute the Rouge-L score for each model-generated summary $s$ and $\mathadsp$. The model-generated summary $s$ is added to $\mathadsp$ only if the Rouge-L score is less than $T_s$.

\textbf{Dialogue Filtering.} During each step of utterance generation, we filter out utterances that are less than 5 tokens in length. When obtained a dialogue $d$,
we compute the semantic embedding of $d$ and $\forall d' \in \mathadp$ using Sentence Transformer \cite{reimers-gurevych-2019-sentence}, namely $e_d$ and $e_{d'}$.
We calculate the cosine similarity between $e_d$ and other embeddings, take the top 5 values and obtain their average value. If the resulting value is less than $T_d$, we add it to the $\mathadp$. Otherwise, we continue with the step of generating utterances. If the number of utterances in a dialogue data $d$ is greater than 10 and still does not meet the requirements, we reset it and regenerate the utterances.

\subsection{Evaluation of Augmented Dialogues: \textsc{SemanticDiversity}}
\label{sec:sd_metric}

In order to evaluate the semantic diversity of the augmented dialogues, we design a metric called \textsc{SemanticDiversity} (SD), as shown in \autoref{alg1}. Given seed data $\mathbf{D}_{seed}$ and augmented data $\mathbf{D}_{aug}$, the output of the algorithm is the semantic diversity value $v$. Firstly, we compute sentence embeddings of seed data to get $\mathbf{H}_{seed}$ and $\mathbf{H}_{aug}$ using Sentence-Transformer\footnote{\url{https://www.sbert.net/}. In this paper, we choose \texttt{all-mpnet-base-v2} as the sentence encoder.} \cite{Reimers2019SentenceBERTSE}. Then we run KMeans algorithm \cite{scikit-learn} on $\mathbf{H}_{seed}$, and the number of clusters was set to $\sqrt{|\mathbf{D}_{seed}|/2}$. Next, we predict the nearest cluster centroid $h_{nearest}$ for every $h_i \in \mathbf{H}_{aug}$, calculate the Euclidean distance $v_i$ between them, and then add $v_i$ to the set $V$. The average score of $V$ is the final semantic diversity value $v$. The larger $v$ is, the sparser the distribution of augmented data in the semantic space, and the more diverse the data is. 

\begin{algorithm}[t]
    \renewcommand{\algorithmicrequire}{\textbf{Input:}}
    \renewcommand{\algorithmicensure}{\textbf{Output:}}
    \caption{\textsc{SemanticDiversity}}
    \label{alg1}
    \begin{algorithmic}[1]
        \Require seed data $\mathbf{D}_{seed}$, augmented data $\mathbf{D}_{aug}$
        \Ensure semantic diversity value $v$
        \State $\mathbf{H}_{seed}, \mathbf{H}_{aug} \leftarrow \text{Encoding}(\mathbf{D}_{seed}, \mathbf{D}_{aug})$ \Comment{Encode seed data and augmented data into hidden states via Sentence-Transformers}
        \State Run KMeans algorithm on $\mathbf{H}_{seed}$ with $\sqrt{|\mathbf{D}_{seed}|/2}$ clusters
        \State $V \leftarrow \{\}$
        \For{$h_i \in \mathbf{H}_{aug}$}
            \State Calculate the Euclidean distance $v_i$ between $h_i$ and its nearest cluster centroid $h_{nearest}$ among the $\sqrt{|\mathbf{D}_{seed}|/2}$ clusters
            \State $V \leftarrow V \cup v_i$
        \EndFor
        \State $v = \text{mean}(V)$ \Comment{Get the average score of $V$}
        \State \Return $v$
    \end{algorithmic}
\end{algorithm}
\section{Experiments}

\subsection{Datasets}

We evaluate various data augmentation methods on \daily \cite{Li2017DailyDialogAM}, a chit-chat dataset that contains high-quality human conversations about daily life. To simulate low-resource scenarios, we randomly sample 100 dialogues for training, 100 for validation, and 1000 for testing, respectively. The training data is used as the seed dataset for subsequent experiments.

\subsection{Baseline Methods}

We compare the proposed method with other baseline methods of data augmentation:

\noindent\textbf{MLM.} Similar to \citet{Cai2020DataMT} and \citet{Ng2020SSMBASM}, we mask 15\% tokens per seed dialogue, and reconstruct these tokens with RoBERTa-Large \cite{Liu2019RoBERTaAR}.

\noindent\textbf{ICL.} First, we sample 5 random dialogues from the seed dataset and concatenate them to construct a prompt. Given the prompt, we use \llm to generate a new dialogue with nucleus sampling decoding ($p=0.9$). The maximum utterance length is 50 and the maximum number of dialogue turns is 10.

\noindent\textbf{ICL$_{context=n}$.} Based on the above ICL, we sample an additional seed dialogue with the first $n$ turns of context to prompt the \llm. In this paper, we set $n$ to 1/2/3, and name the corresponding methods as ICL$_{context=1}$, ICL$_{context=2}$, and ICL$_{context=3}$, respectively.

\subsection{Implementation Settings}

We set hyper-parameters for the three steps of our method according to the performance on the validation data. For Seed Dialogue Summarization (\autoref{sec:summarization}), we use beam-search decoding with $beam\_size = 3$. For Dialogue Summary Augmentation (\autoref{sec:dialogue_summary_generation}), we use nucleus sampling decoding with $p = 0.9$ and $temperature = 0.9$ for generating more diverse dialogues with LLM. The hyper-parameters of Dialogue Generation with Summary (\autoref{sec:dialogue_data_generation}) are similar to Dialogue Summary Augmentation but $temperature = 0.6$ for better dialogue fluency. For Data Filtering (\autoref{sec:data_filtering}), $T_s$ is set to 0.35, while $T_d$ is set to 0.8.

Given the \seedd, we collect 1,000 dialogues for each augmentation method. After obtaining the augmented dataset, we fine-tune a pre-trained encoder-decoder model, BART-large \cite{Lewis2019BARTDS}, with a learning rate of 5e-5, a batch size of 32, and the maximum sequence length of 512. We adopt the checkpoint with the lowest validation set loss for evaluation. During the inference stage, we use greedy search decoding and limited the maximum decoding length to 50.

\subsection{Evaluation Metrics}

\label{sec:metrics}

\noindent\textbf{Automatic Evaluation.} We use \llm to compute the average perplexity (PPL) to evaluate the data fluency. For the data diversity, we employ Distinct-1/2 \cite{Li2015ADO} (Dist-1 and Dist-2) for word-level evaluation and \textsc{SemanticDiversity} (described in \autoref{sec:sd_metric}) for semantic-level evaluation.


For model prediction, we use SacreBLEU \cite{post-2018-call} and Rouge-L \cite{lin-2004-rouge} to measure the similarity of the predicted response to the ground truth, and corpus-level Distinct-1/2 \cite{Li2015ADO} to measure the text diversity.

\noindent\textbf{Human Evaluation.} For model prediction, we randomly select 50 dialogue context-response pairs respectively. Three annotators are asked to rate the response quality from three aspects: (1) \textbf{Fluency}: whether the response is smooth and grammatically correct. (2) \textbf{Coherence}: whether the response is coherent with the context. (3) \textbf{Informativeness}: whether the response is informative or not. The rating range is [0, 1, 2], where a higher score indicates better quality. Model prediction's final score is averaged across the three annotators.
\section{Result and Analysis}

\subsection{Evaluating Augmented Dialogue}

\begin{table}[h]
    \centering
    \begin{tabular}{lcccc}
        \toprule
        Methods & PPL & Dist-1 & Dist-2 & SD \\
        \midrule
        MLM & 6.77 & 1.76 & 7.01 & 61.81 \\
        ICL & 3.81 & 3.42 & 21.47 & 75.85 \\
        ICL$_{context=1}$ & 4.00 & 3.18 & 19.57 & 73.49 \\
        ICL$_{context=2}$ & 4.26 & 3.03 & 18.41 & 72.37 \\
        ICL$_{context=3}$ & 4.46 & 2.83 & 16.91 & 71.32 \\
        \midrule
        SDA & 3.58 & 3.01 & 16.45 & 77.52 \\
        \;w/o SF & 6.01 & 3.87 & 22.35 & 69.02 \\
        \;w/o DF & 5.93 & 4.10 & 22.80 & 69.82 \\
        \;w/o SF+DF & 5.80 & 3.97 & 21.98 & 68.97 \\
        
        \bottomrule
    \end{tabular}
    \caption{Automatic evaluation on augmented dialogue, along with the ablation results. SD refers to \textsc{SemanticDiversity}.}
    \label{tab:auto_data}
\end{table}

\begin{figure}
    \centering
    \includegraphics[width=\linewidth]{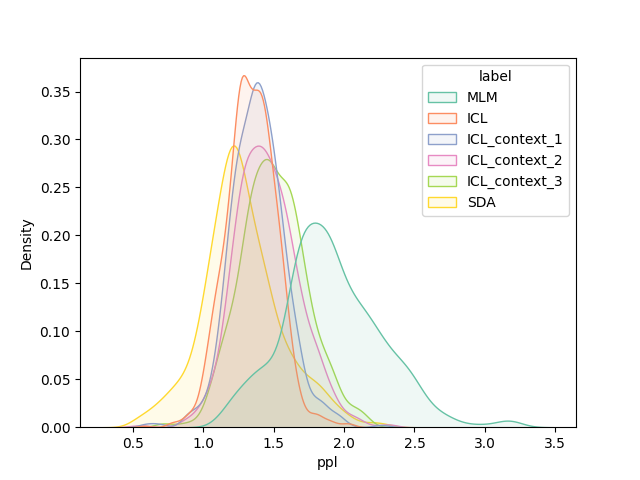}
    \caption{Dialogue perplexity distribution with different data augmentation methods (best viewed in color).}
    \label{fig:ppl_distribution}
\end{figure}

\begin{figure}
    \centering
    \includegraphics[width=\linewidth]{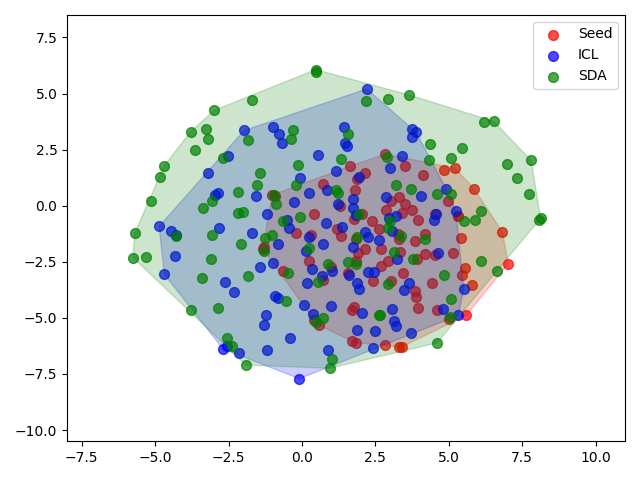}
    \caption{The t-SNE visualization of augmented dialogues.}
    \label{fig:tsne}
\end{figure}


\begin{table}[t]
    \centering
    \begin{tabularx}{\linewidth}{Xcccc}
        \toprule
        Methods & B. & R. & D-1 & D-2   \\
        \midrule
        $\varnothing$  & 0.87 & 9.47 & 2.27 & 10.13  \\
        +MLM & 0.94 & 9.78 & 2.15 & 9.01  \\
        +ICL & 1.32 & 12.61 & 3.73 & 16.48  \\
        +ICL$_{context=1}$ & 1.23 & 12.54 & 3.65 & 15.21 \\
        +ICL$_{context=2}$ & 1.08 & 11.82 & 3.31 & 13.44  \\
        +ICL$_{context=3}$ & 0.98 & 10.45 & 2.7 & 11.63 \\
        \midrule
        +SDA & 1.34 & 12.96 & 4.09 & 18.56 \\
        \;\;w/o SF & 1.15 & 11.93 & 3.21 & 15.16 \\
        \;\;w/o DF & 0.99 & 11.79 & 3.53 & 14.66  \\
        \;\;w/o SF+DF & 1.00 & 11.04 & 3.30 & 14.54  \\
        \bottomrule
    \end{tabularx}
    \caption{Automatic evaluation on dialogue model prediction, along with the ablation results. $\varnothing$ only uses the seed data to fine-tune the dialogue model. B./R./D-1/D-2 stands for SacreBLEU/ROUGE-L/Dist-1/Dist-2 respectively.}
    \label{tab:auto_model}
\end{table}

\begin{table}[h]
    \centering
    \begin{tabular}{lcccc}
        \toprule
        Dataset & Flu. & Coh. & Inf. & Average  \\
        \midrule
        ICL & 1.51 & 1.05 & 1.04 & 1.20 \\
        ICL$_{context=1}$ & 1.53 & 1.08 & 0.98 & 1.19 \\
        ICL$_{context=2}$ & 1.48 & 0.91 & 0.84 & 1.08  \\
        ICL$_{context=3}$ & 1.44 & 0.74 & 0.82 & 1.00  \\
        SDA & 1.62 & 1.20 & 1.19 & 1.34  \\

        \bottomrule
    \end{tabular}
    \caption{Human evaluation on dialogue model prediction with different data augmentation methods. Flu./Coh./Inf. stands for Fluency/Coherence/Informativeness respectively.}
    \label{tab:human_model}
\end{table}

\begin{figure*}
    \centering
    \includegraphics[width=\linewidth]{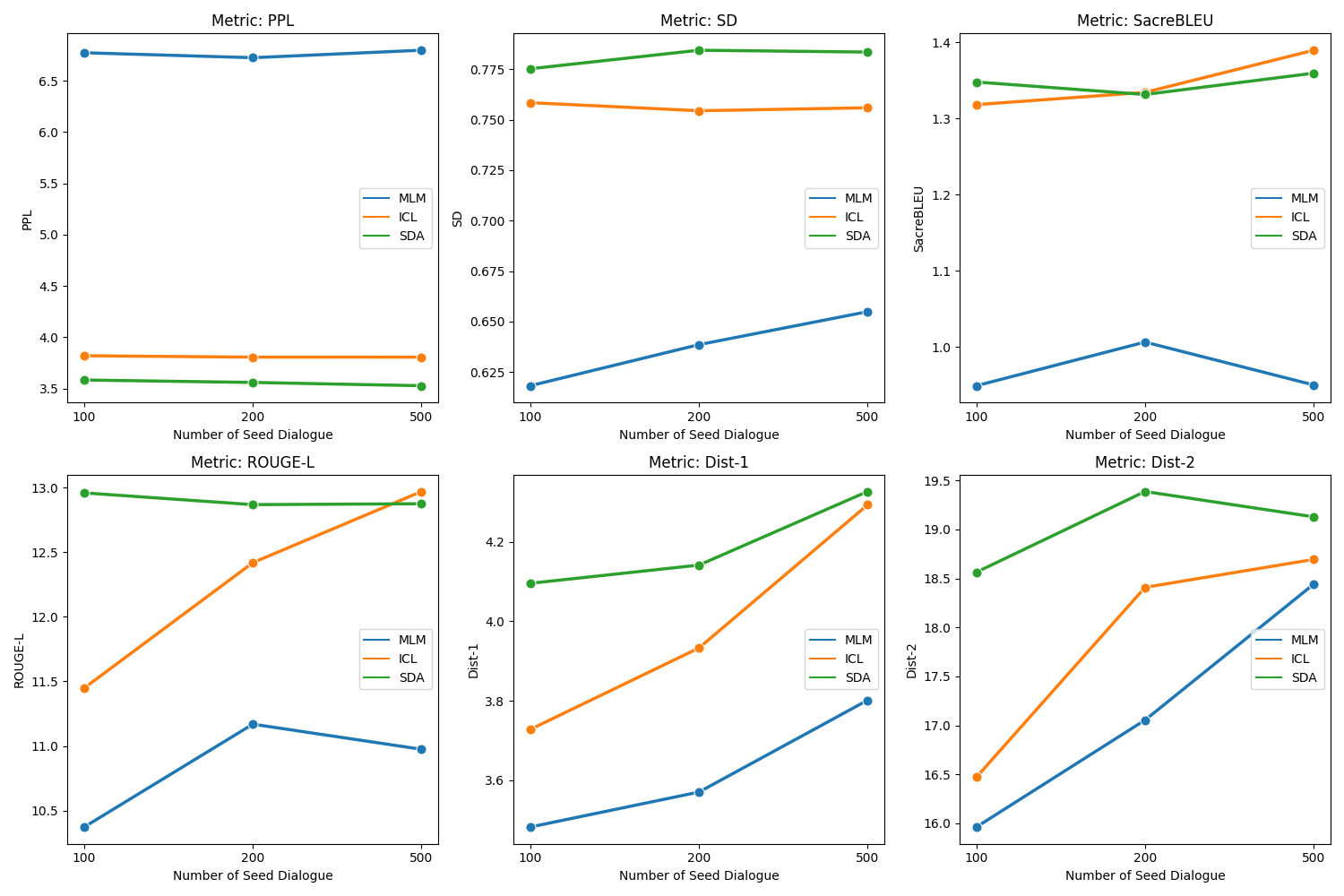}
    \caption{The performance of various data augmentation methods given 100/200/500 seed dialogues (best viewed in color). The initial two (PPL, SD) are metrics for evaluating the augmented data, while the latter four  are metrics for evaluating the predictions of the model trained on the augmented data.}
    \label{fig:ablation_num_seed}
\end{figure*}

To begin with, it is primary to evaluate the augmented dialogue generated by different methods. 
The result of the automatic evaluation is summarized in \autoref{tab:auto_data} and \autoref{fig:ppl_distribution}, which demonstrates that our SDA-generated augmented dialogue data with the lowest perplexity, indicating the highest level of fluency in the text. MLM got the highest perplexity, indicating that the mask-then-reconstruct approach cannot achieve the fluency of seed data. It is worth noting that although the Dist-1/2 scores of SDA are lower than ICL, ICL$_{context=1}$ and ICL$_{context=2}$, the \textsc{SemanticDiversity} of SDA is the highest. In other words, the diversity of SDA is not significant at the word-level, but it performs best at the semantic-level. For the baselines based on ICL, the less the number of contexts used, the higher the Dist-1/2 and SD. 

In addition to the \textsc{SemanticDiversity} value showing the semantic diversity of the augmented dialogues, we conduct t-SNE visualization for both ICL and SDA methods. We sample 100 dialogues from the augmented data obtained from ICL and SDA respectively, then use Sentence Transformer to calculate their sentence embeddings, and finally perform t-SNE visualization, as shown in the \autoref{fig:tsne}. We observe that:
\begin{itemize}
\item SDA demonstrates higher diversity compared to ICL, aligning with the \textsc{SemanticDiversity} values presented in \autoref{tab:auto_data}. This underscores the efficacy of the \textsc{SemanticDiversity} metric.
\item ICL exhibits some distribution shift compared to the seed data, while the SDA fully covers the distribution of the seed data. This indicates that SDA has better controllability than ICL.
\end{itemize}

\subsection{Evaluating Dialogue Model}
\label{evaluating_dialog_model}

After determining that our augmented dialogues are of fairly high quality and diversity, we attempt to use the augmented datasets as training data for the dialogue model. The experimental results with automatic evaluations are summarized in \autoref{tab:auto_model}, which indicates that SDA outperforms all the baselines on all the automatic metrics. This confirms the effectiveness of our dialogue augmentation method, which could generate high-quality and semantic-diverse dialogue data. We can further observe that: (1) The data quality produced by the MLM method is unsatisfactory. Consequently, the Dist-1/2 of model predictions are inferior to those obtained by training solely on seed data. (2) The ICL method performs well on the model, demonstrating large models can generate high-quality dialogue data. However, as more rounds of context are given, the diversity of augmented data diminishes, leading to a gradual decline in the model's performance. In addition, based on the results from \autoref{tab:auto_data} and \autoref{tab:auto_model}, we observe that there is a positive correlation between the SD of a dataset and the performance of models trained using this dataset. This suggests that the SD metric provides an effective measure for evaluating the diversity of augmented datasets.

The human evaluation results are presented in Table \ref{tab:human_model}. As shown in the table, our proposed method outperforms other data augmentation methods on all three criteria, achieving an average score of 1.34. The performance of the ICL method are sub-optimal, ranking second to our method. However, adding the contextualized ICL datasets (ICL$_{context=1}$, ICL$_{context=2}$, and ICL$_{context=3}$) does not lead to consistent improvement, and the best performance is achieved using our proposed method. In addition, we find that these methods have little difference in fluency, indicating that the pre-trained model has strong generation capabilities and is not greatly affected by the dataset. However, there is a large gap in coherence and informativeness, which is highly related to the relevance of the data. In summary, these findings confirm the effectiveness of our method in generating more fluent, coherent, and informative responses in open-domain dialogue generation.

\subsection{Ablation Analysis of Filtering}

We further explore the necessity of the data filtering module by ablation experiments. Specifically, we compare SDA with three variants of it: (1) SDA without Summary Filtering (w/o SF), (2) SDA without Dialogue Filtering (w/o DF), (3) SDA without Summary Filtering and Dialogue Filtering (w/o SF+DF). The ablation results are shown in \autoref{tab:auto_data} and \autoref{tab:auto_model}. From these results, we find that: 

\begin{itemize}
\item Without data filtering, both fluency and semantic diversity show a significant decrease, although data Dist-1/2 improves.
\item For model predictions, all three variants show significant decreases in each metric. Among them, SDA w/o SF+DF has the lowest average score.
\end{itemize}

The above findings indicate that both SF and DF are indispensable for our method.

\subsection{Ablation Analysis of Various Number of Seed Dialogue}

We also conduct experiments to evaluate the performance of various data augmentation methods given a varying number of seed dialogues. Specifically, the number of seed dialogues selected is 100, 200, and 500 respectively. The data augmentation methods chosen for comparison include MLM, ICL, and SDA. The detailed results are illustrated in \autoref{fig:ablation_num_seed}. We observe that:

\begin{itemize}
\item With varying number of seed data, the augmented data generated through the SDA method exhibits superior fluency.
\item As the quantity of seed data increases, the diversity metrics (SD, Dist-1, Dist-2) for all augmentation methods show improvement.
\item When the number of seed data is relatively small, the advantages of SDA over other methods are more pronounced.
\end{itemize}







\section{Conclusion}

This paper presents SDA, a data augmentation approach for low-resource open-domain dialogue generation. Our approach improves the controllability of \llm by taking dialogue summary as planning, which generates high-quality and diverse dialogue data without distribution shift compared to the seed data. In order to evaluate data diversity at the semantic-level, we design a metric, \textsc{SemanticDiversity}, instead of word-level which is often used in the previous studies. Experimental results show that SDA can augment high-quality dialogue with diverse semantics, which can be further used to improve model performance in low-resource scenarios. Furthermore, \textsc{SemanticDiversity} metric exhibits a strong positive correlation with the performance of the dialogue model.


\section*{Limitations}

In this paper, we develop a simple open-domain dialogue augmentation method with \llm. Our method strongly relies on the ICL capacity of \llm, which is related to the scale of the model \cite{Kaplan2020ScalingLF, Brown2020LanguageMA}. However, due to GPU resource limitations, we have not performed any experiments with larger scale \llm. In general, the larger the model is, the better the augmented dialogue is. Moreover, we have not explored the upper limit of the number of augmented data given the seed data. When the size of the augmented data grows to a certain level, simply boosting the number of data often becomes less efficient for model performance. This paper applied the method solely to the \daily dataset. For conducting the method in other dialogue scenarios, it may require modifying the instructions to ensure that \llm generates dialogue data that aligns more with the expectations.
\section*{Ethics Statement}

Due to the training data and training methods of \llm, there is a potential risk of generating biased, harmful, or otherwise unwanted output. More fine-grained analysis and filtering work needs to be done before the actual application.

\bibliography{custom}

\appendix

\section{Dialogue Augmentation Examples}
\label{sec:examples}
We present the examples of dialogue augmentation in \autoref{tab:examples}.

\begin{table*}[!ht]
    \centering
    \begin{tabular}{cc}
        \toprule
        \multicolumn{2}{c}{\centering Seed Dialogue} \\
        \hline
        \multicolumn{2}{l}{\raggedright User A: I'm going to Japan this year on vacation.} \\
        \multicolumn{2}{l}{\raggedright User B: Have you ever been to America?} \\
        \multicolumn{2}{l}{\raggedright User A: No, but I'd really like to.} \\
        \multicolumn{2}{l}{\raggedright User B: You'd like it.} \\
        \midrule

        \multicolumn{2}{c}{\centering MLM} \\
        \hline
        \multicolumn{2}{l}{\raggedright User A: I'm going to Japan this year on vacation.} \\
        \multicolumn{2}{l}{\raggedright User B: Have you ever been to \textcolor{blue}{Hawaii}?} \\
        \multicolumn{2}{l}{\raggedright User A: No, but I'd really like to.} \\
        \multicolumn{2}{l}{\raggedright User B: You'd like it.} \\
        \midrule

        \multicolumn{2}{c}{\centering ICL$_{context=1}$} \\
        \hline
        \multicolumn{2}{l}{\raggedright User A: I'm going to Japan this year on vacation.} \\
        \multicolumn{2}{l}{\raggedright User B: \textcolor{blue}{Where do you plan to go?}} \\
        \multicolumn{2}{l}{\raggedright User A: \textcolor{blue}{I am thinking about going to Mount Fuji.}} \\
        \multicolumn{2}{l}{\raggedright User B: \textcolor{blue}{Did you go last year?}} \\
        \midrule

        \multicolumn{2}{c}{\centering ICL$_{context=2}$} \\
        \hline
        \multicolumn{2}{l}{\raggedright User A: I'm going to Japan this year on vacation.} \\
        \multicolumn{2}{l}{\raggedright User B: Have you ever been to America?} \\
        \multicolumn{2}{l}{\raggedright User A: \textcolor{blue}{I went to Florida a few years ago.}} \\
        \multicolumn{2}{l}{\raggedright User B: \textcolor{blue}{that must have been nice!}} \\
        \midrule

        \multicolumn{2}{c}{\centering ICL$_{context=3}$} \\
        \hline
        \multicolumn{2}{l}{\raggedright User A: I'm going to Japan this year on vacation.} \\
        \multicolumn{2}{l}{\raggedright User B: Have you ever been to America?} \\
        \multicolumn{2}{l}{\raggedright User A: No, but I'd really like to.} \\
        \multicolumn{2}{l}{\raggedright User B: \textcolor{blue}{Have you got your American visa yet?}} \\
        \midrule

        \multicolumn{2}{c}{\centering SDA} \\
        \hline
        \multicolumn{2}{l}{\raggedright User A: \textcolor{blue}{How do you think we should prepare for our presentation?}} \\
        \multicolumn{2}{l}{\raggedright User B: \textcolor{blue}{I think we should practice about 3 times a week.}} \\
        \multicolumn{2}{l}{\raggedright User A: \textcolor{blue}{That's a good idea. But do you think we should practice in front of the mirror?}} \\
        \multicolumn{2}{l}{\raggedright User B: \textcolor{blue}{I think we should. You can see yourself and correct your mistakes.}} \\

        \bottomrule
    \end{tabular}
    \caption{Examples of augmented dialogue.}
    \label{tab:examples}
\end{table*}

\end{document}